\documentclass[letterpaper,10pt,conference]{ieeeconf}
\IEEEoverridecommandlockouts
\usepackage{iftex}
\ifPDFTeX
\else
  \AtBeginDvi{}
\fi
\ifPDFTeX
\else
  \usepackage{fontspec}
\fi
\usepackage{graphicx}
\usepackage{amsmath,amssymb}
\usepackage{booktabs}
\usepackage{multirow}
\usepackage{xcolor}
\usepackage{url}
\newcommand{\ilarevision}[1]{{#1}}
\newcommand{\WideTableStyle}{\centering\small\renewcommand{\arraystretch}{1.1}}
\newcommand{\CompactTableStyle}{\centering\footnotesize\renewcommand{\arraystretch}{1.1}}

\makeatletter
\newcommand{\FirstPageFigure}[1]{%
  \begin{minipage}{\textwidth}
  \centering
  \includegraphics[width=0.88\textwidth]{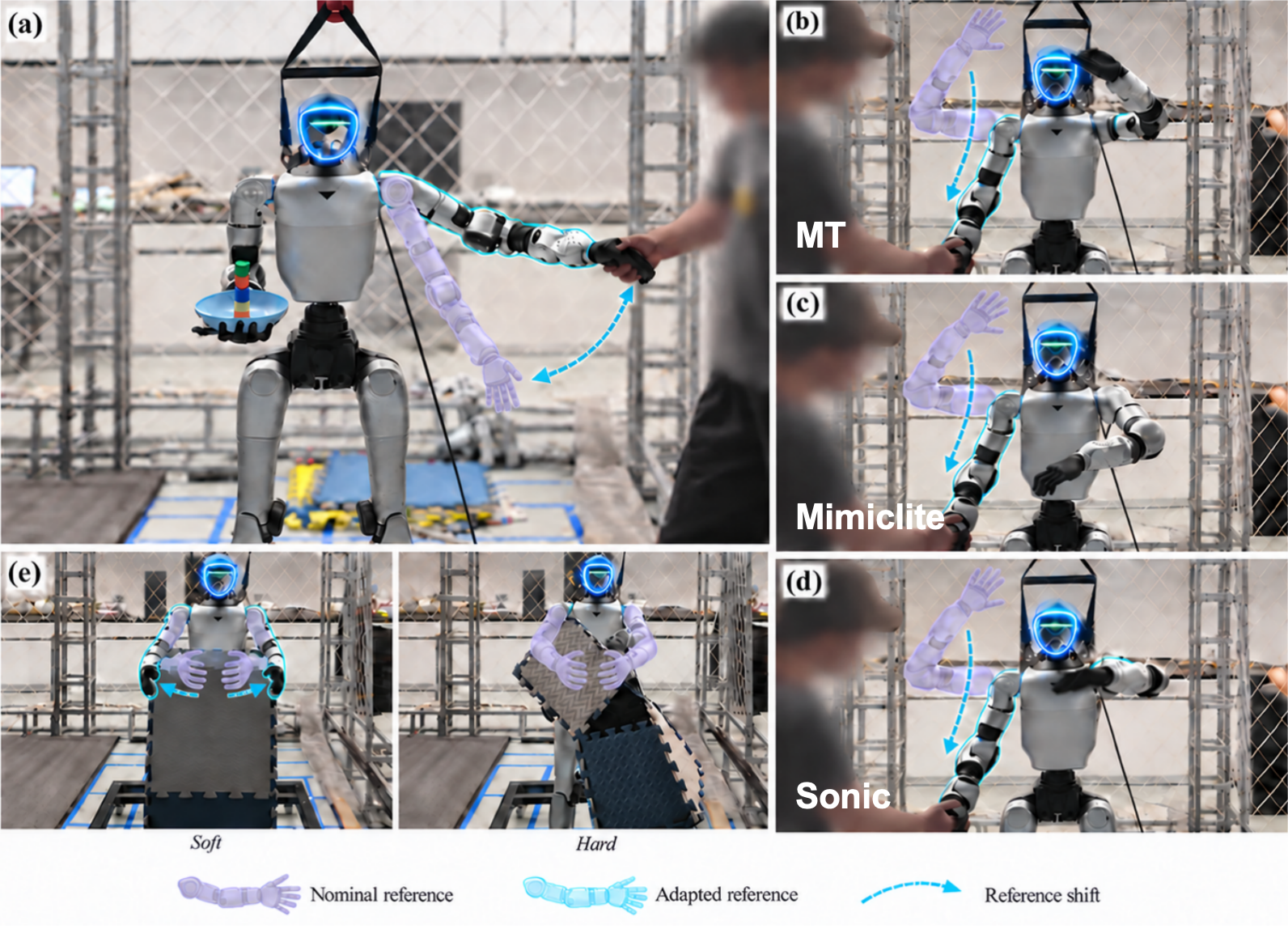}
  \def\@captype{figure}%
  \caption{#1}\label{fig:teaser}
  \end{minipage}\vspace{8pt}%
}
\makeatother

\title{\LARGE\bfseries ILA: A Transferable Interface-Level Adapter \\for Compliant Humanoid Motion Tracking}

\author{Zhuochang Cai$^{1,*}$, Zihao Yan$^{1,*}$, Jinxuan Zhu$^{2,*}$, Yanjie Li$^{1}$, and Hao Xiong$^{1,\dagger}$%
\thanks{$^{*}$These authors contributed equally to this work and share first authorship.}%
\thanks{$^{1}$Harbin Institute of Technology, Shenzhen, China.}%
\thanks{$^{2}$National University of Singapore, Singapore.}%
\thanks{$^{\dagger}$Corresponding author: Hao Xiong.}%
\thanks{This work has been submitted to the IEEE for possible publication. Copyright may be transferred without notice, after which this version may no longer be accessible.}%
}
\IEEEaftertitletext{\FirstPageFigure{ILA adds physical guidance through reference adaptation while preserving pretrained whole-body policies. (a) Human-guided reference adaptation. (b--d) The same ILA trained with standard MotionTracking shifts a nominal raised-arm reference downward on MotionTracking (MT), MimicLite, and Sonic, with both adapter and tracker parameters fixed. (e) Soft and hard box interaction: the box remains assembled in the soft example and comes apart in the hard example. In (a--d), lavender and cyan indicate conceptual nominal and adapted references; overlays are illustrative, not measured trajectories.}
}
\begin{document}
\maketitle
\newcommand{\MTOffTarget}{17.801}
\newcommand{\ILATarget}{5.684}
\newcommand{\TargetReductionPct}{68.1\%}
\newcommand{\MTOffWork}{13.675}
\newcommand{\ILAWork}{10.859}
\newcommand{\MTOffRoot}{1.642}
\newcommand{\ILARoot}{1.246}
\newcommand{\MTOffFall}{2.881}
\newcommand{\ILAFall}{0.000}
\newcommand{\MTOffInactive}{2.749}
\newcommand{\ILAInactive}{2.815}
\newcommand{\InactiveChange}{0.066}
\newcommand{\MLBefore}{10.182}
\newcommand{\MLAfter}{3.705}
\newcommand{\MLReductionPct}{63.6\%}
\newcommand{\SonicBefore}{14.871}
\newcommand{\SonicAfter}{3.663}
\newcommand{\SonicReductionPct}{75.4\%}
\newcommand{\DirectTarget}{14.83$\pm$0.08}
\newcommand{\IKTarget}{11.86$\pm$0.12}
\newcommand{\MRTarget}{7.99$\pm$0.32}
\newcommand{\MRvsDirect}{$-$6.84}
\newcommand{\MRvsIK}{$-$3.86}
\newcommand{\SeedMRvsIK}{$-$4.39 [$-$4.70, $-$4.09]}
\newcommand{\LCBTeacher}{5.326}
\newcommand{\LCBStudent}{5.684}
\newcommand{\LCBCross}{6.838}
\newcommand{\LCBCurrent}{7.538}
\newcommand{\LCBMean}{17.678}
\newcommand{\LCBShift}{17.740}
\newcommand{\GOneDropCI}{$-$12.1 [$-$14.1, $-$10.1]}
\newcommand{\MLDropCI}{$-$6.48 [$-$7.62, $-$5.32]}
\newcommand{\SonicDropCI}{$-$11.2 [$-$13.2, $-$9.29]}
\newcommand{\LCBGapTeacher}{0.358 [0.287, 0.434]}
\newcommand{\LCBvsCross}{$-$1.15 [$-$1.53, $-$0.81]}
\newcommand{\LCBvsCurrent}{$-$1.85 [$-$2.39, $-$1.34]}
\newcommand{\LCBvsMean}{$-$11.99 [$-$13.16, $-$10.85]}
\newcommand{\LCBvsShift}{$-$12.06 [$-$13.22, $-$10.90]}
\newcommand{\WhenYield}{0.145 [0.109, 0.183]}
\newcommand{\WhenHold}{0.254 [0.207, 0.305]}
\newcommand{\WhereDelta}{0.513 [0.466, 0.558]}
\newcommand{\HowMuchRho}{0.712 [0.626, 0.793]}
\newcommand{\NoForceJoint}{0.320}
\newcommand{\NoForceKP}{8.571}
\newcommand{\NoForceRange}{0.31--1.01}
\newcommand{\StabTarget}{7.598$\pm$0.368}
\newcommand{\StabCV}{4.8\%}
\newcommand{\StabFallRange}{2.92--3.53\%}
\newcommand{\FightFall}{11.3}
\newcommand{\FightRec}{4.59}

\newcommand{\HWMTPullBase}{\textit{NR}@74.48}
\newcommand{\HWMTPullILA}{16.88}
\newcommand{\HWMLPullBase}{\textit{NR}@69.64}
\newcommand{\HWMLPullILA}{10.72}
\newcommand{\HWSonicPullBase}{\textit{NR}@77.42}
\newcommand{\HWSonicPullILA}{18.64}
\newcommand{\HWGentlePull}{11.73}

\newcommand{\HWMTBaseDisturb}{0/10}
\newcommand{\HWMTILADisturb}{10/10}
\newcommand{\HWMTBaseHandshake}{0/10}
\newcommand{\HWMTILAHandshake}{10/10}

\newcommand{\HWMLBaseDisturb}{0/10}
\newcommand{\HWMLILADisturb}{10/10}
\newcommand{\HWMLBaseHandshake}{0/10}
\newcommand{\HWMLILAHandshake}{10/10}

\newcommand{\HWSonicBaseDisturb}{0/10}
\newcommand{\HWSonicILADisturb}{10/10}
\newcommand{\HWSonicBaseHandshake}{0/10}
\newcommand{\HWSonicILAHandshake}{10/10}

\newcommand{\HWGentleDisturb}{10/10}
\newcommand{\HWGentleHandshake}{10/10}

\newcommand{\HWMTCompPull}{18.47}
\newcommand{\HWMTCompDisturb}{10/10}
\newcommand{\HWMTCompHandshake}{10/10}

\begin{abstract}
\ilarevision{
Humanoid motion tracking is fundamental to reproducing human motions on humanoid robots and serves as an important tool for teleoperation and motion data collection. However, directly tracking reference motions can be unsafe in constrained environments or during physical human–robot interaction, motivating the need for compliant motion tracking. We present \textbf{ILA}, an \textbf{I}nterface-\textbf{L}evel \textbf{A}dapter for compliant humanoid motion tracking without dedicated interaction sensors. ILA adapts motion references from the robot’s observation history, allowing the robot to respond to physical guidance while preserving the capabilities of a pretrained motion tracker. As an interface-level module, ILA can be directly plugged into different pretrained motion trackers without retraining either the adapter or the tracking policies. ILA is trained entirely in simulation to predict reference corrections induced by physical interaction. We evaluate ILA with three humanoid motion trackers and achieve a 63.6–75.4\% reduction in guidance-target error compared with their original policies. Matched simulation studies and hardware experiments further show that ILA improves responsiveness to human guidance while preserving nominal motion-tracking performance. These results demonstrate compliant guidance as a reusable, sensor-free extension to pretrained humanoid motion trackers.
}
\end{abstract}

\section{Introduction}
\ilarevision{
Humanoid motion tracking brings diverse human motions to robots, supporting teleoperation and motion data collection~\cite{luo2026sonic,chen2025gmt,he2024h2o,he2024omnih2o}. During physical interaction, faithfully following a prescribed motion can conflict with yielding to human guidance. Compliant tracking should accommodate guidance while retaining existing motion skills (Fig.~\ref{fig:teaser}).
}

\ilarevision{
Learning-based compliance typically requires retraining or redesigning the motion controller~\cite{lu2025gentle,margolis2025softmimic,chen2025chip}. Analytical retrofits rely on explicit wrench estimation and prescribed force--motion relations~\cite{shi2026mcc}. We seek reusable compliance for pretrained trackers without changing their architecture or parameters or adding dedicated interaction sensors.
}

\ilarevision{
We present ILA, an Interface-Level Adapter that predicts reference corrections from the nominal reference and the robot's observation history. The pretrained tracker follows the adapted reference and handles whole-body execution and stabilization. Its architecture and parameters remain unchanged, which we term \emph{policy-preserving compliance retrofit}. ILA is trained entirely in simulation using paired nominal and modified references. The trained adapter can be reused with other trackers through deterministic mappings of observations and motion references.
}

\ilarevision{
The same trained adapter reduces guidance target error by 63.6--75.4\% across three independently trained trackers, without finetuning either the adapter or the tracking policies. Matched simulation comparisons assess guidance gains alongside nominal-motion preservation. Hardware experiments demonstrate human-guided motion with concurrent task execution and whole-body coordination.
}

Our contributions are:
\begin{itemize}
\item \ilarevision{
\textbf{Policy-preserving compliance retrofit.} Compared with policy-training approaches and wrench-based retrofits, ILA adds physical guidance without tracker retraining or explicit wrench estimation. It learns reference corrections from causal interaction history, retaining the pretrained policy for whole-body execution and stabilization.
}
\item \ilarevision{
\textbf{Cross-tracker reuse of learned compliance.} ILA transfers the same learned adapter across independently trained motion trackers through deterministic observation and reference mappings. Both adapter and tracker parameters remain fixed, avoiding per-tracker compliance training.
}
\item \ilarevision{
\textbf{Systematic evaluation of compliance and motion preservation.} Matched comparisons and hardware experiments evaluate guidance gains together with nominal-motion tracking, spatial selectivity, and post-release recovery, including comparisons with direct policy finetuning.
}
\end{itemize}

\begin{figure*}[t]
\centering
\includegraphics[width=0.96\textwidth]{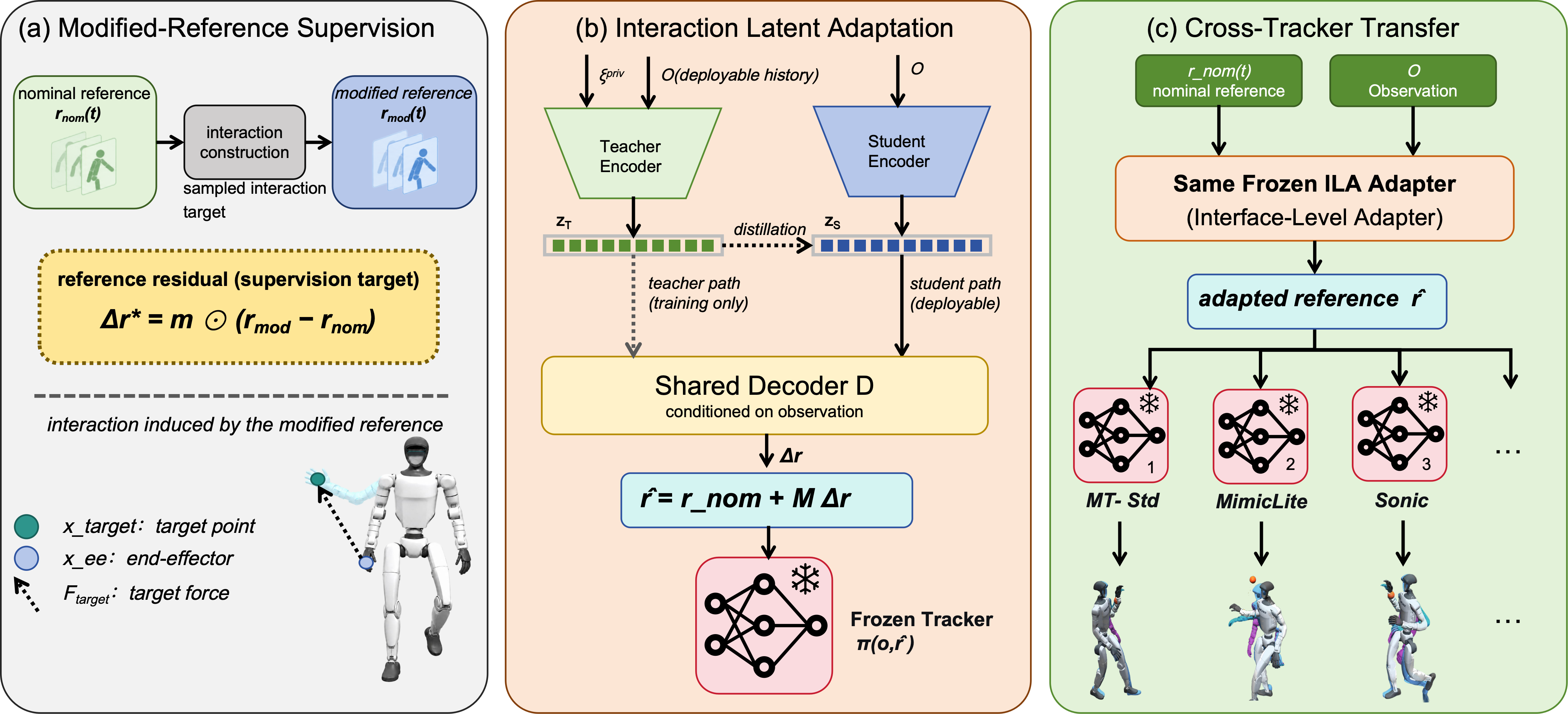}
\caption{ILA adapts the reference while preserving the controller. (a) Nominal--modified reference pairs supervise the residual $\Delta r^*=m\odot(r^{\mathrm{mod}}-r^{\mathrm{nom}})$. (b) $O$ denotes the observation history and nominal reference $(h_t,r_t)$. A privileged teacher and a causal-history student share a decoder; at deployment, the student adapts motion intent and the frozen tracker performs whole-body execution and stabilization. (c) The same frozen adapter is reused with other compatible frozen trackers through deterministic observation and reference mappings.}
\label{fig:overview}
\end{figure*}

\section{Related Work}
\subsection{\ilarevision{Reusable humanoid motion control}}
Motion tracking has evolved from single-clip imitation~\cite{peng2018deepmimic} to generalist whole-body controllers trained on large motion corpora~\cite{mahmood2019amass,luo2023phc,luo2023pacer,chen2025gmt,luo2026sonic,tessler2024maskedmimic}. Teleoperation and sim-to-real methods deploy these capabilities on hardware~\cite{he2024h2o,he2024omnih2o,he2024hover,he2025asap,gu2024humanoidgym,fu2025humanplus,ji2024exbody2}, while BeyondMimic uses guided diffusion to extend motion tracking to unseen tasks~\cite{liao2025beyondmimic}. \ilarevision{
However, closely following a prescribed motion can conflict with yielding to human guidance.
}

\subsection{\ilarevision{Learned and analytical compliance}}
Classical impedance and admittance control specify force--motion relations~\cite{hogan1985impedance,ott2015cartesian,ott2010unified,ott2008passivity}. Learned approaches reproduce virtual mass--spring--damper responses~\cite{xu2025facet}, unify force and position control using interaction history~\cite{zhi2025unifp}, or maintain loco-manipulation accuracy under external forces~\cite{zhang2026falcon}. GentleHumanoid learns spring-inspired upper-body compliance~\cite{lu2025gentle}; SoftMimic learns from synthesized compliant trajectories~\cite{margolis2025softmimic}. CHIP modifies observed tracking goals through hindsight perturbation while retaining the original reference for reward computation~\cite{chen2025chip}. Its plug-and-play property concerns integration into tracker training. MotionTracking's compliance configuration is a related policy-level baseline~\cite{axellwppr_motiontracking}. LAC learns commanded linear and angular compliance through augmented motions and teacher--student training~\cite{liu2026lac}, while CEER develops a compliant end-effector--root control interface~\cite{luo2026ceer}. Minimalist Compliance Control (MCC) provides a model-based reference retrofit: it explicitly estimates external wrench from actuator signals and robot kinematics, then determines reference displacement through a prescribed task-space admittance law and inverse kinematics~\cite{shi2026mcc}. \ilarevision{
Learning-based approaches typically incorporate compliance into policy training, while analytical retrofits such as MCC rely on explicit wrench estimation and prescribed admittance dynamics.
}

\subsection{\ilarevision{Adaptation around pretrained policies}}
Robust deployment of learned models motivates adaptation~\cite{chen2026dwp,miki2022perceptive}. Privileged-to-history adaptation is established by RMA and its bipedal extensions~\cite{kumar2021rma,kumar2022arma}. Residual RL adds learned corrections to controller outputs~\cite{johannink2019residual,silver2018residual}. ResMimic uses a frozen general motion tracker and learns task residuals in action space~\cite{zhao2025resmimic}, while SplitAdapter retains a pretrained policy and modulates internal features through context encoders and hierarchical FiLM~\cite{kang2026splitadapter}. Any2Track uses history-informed adaptation to maintain tracking under terrain, force, and dynamics disturbances~\cite{zhang2025any2track}. \ilarevision{
ILA instead learns interaction-dependent reference corrections, enabling compliant guidance while leaving the tracking policy's architecture and parameters unchanged.
}

\section{Method}
\ilarevision{
ILA learns reference corrections from interaction-consistent targets and deployable observation histories, while a fixed tracker executes the adapted motion (Fig.~\ref{fig:overview}).

\subsection{Interface-Level Compliance Adaptation}
ILA makes yielding to guidance compatible with tracking by adapting the nominal reference:
\begin{equation}
\begin{aligned}
 \Delta r_t&=A_\phi(h_t,r_t),\qquad \hat r_t=r_t+M\Delta r_t,\\
 a_t&=\pi_\theta(o_t,\hat r_t),\qquad \theta\ \text{fixed}.
\end{aligned}
\label{eq:adapter}
\end{equation}
Here, $o_t$ is the tracker's native observation, $r_t$ is the nominal reference, and $h_t$ contains causal histories of proprioception and prior tracker actions in ILA's training coordinates. Tracker architecture and parameters remain unchanged.

The adapter predicts an $H$-step residual over 14 arm joints to represent the reference change over time. The mask $M$ inserts these corrections while retaining nominal trunk and leg references; the tracker coordinates whole-body execution through its existing action space. Zero correction restores the nominal reference supplied to the tracker.

For cross-tracker reuse, deterministic mappings align observations and references with ILA's training representation and insert corrections into the target tracker's reference. These mappings account for joint ordering, coordinate conventions, and reference timing. Adapter and tracker parameters remain fixed during transfer evaluation.

\subsection{Interaction-Consistent Reference Supervision}
\label{sec:modref}
A nominal--modified reference pair defines physical guidance and correction targets. The modified reference encodes yielding, holding, and recovery relative to nominal motion.

Nominal motions come primarily from LAFAN1~\cite{harvey2020robust}, AMASS~\cite{mahmood2019amass}, and 100STYLE~\cite{mason2022local}. Following existing interaction-construction procedures~\cite{axellwppr_motiontracking,lu2025gentle}, we sample a finite interaction interval and a temporary upper-body configuration. Selected arm joints move smoothly toward this configuration, hold, and return to the nominal motion; references coincide outside the interval. The environment applies a target-spring wrench computed from the relative hand positions of the nominal and modified clips, tying the training guidance to the desired reference change.

For an $H$-step window, the correction target is
\begin{equation}
 \Delta r^*_{t:t+H-1}
 =m_{t:t+H-1}\odot
 \bigl(r^{\mathrm{mod,u}}_{t:t+H-1}-r^{\mathrm{nom,u}}_{t:t+H-1}\bigr),
\label{eq:target}
\end{equation}
where $\mathrm{u}$ denotes arm joints and $m$ is the interaction mask. Targets follow the interaction offset throughout recovery and specify zero correction outside interaction.

The reference pair defining the wrench and target remains unchanged as ILA adjusts the tracker's input. Tracker actions affect subsequent observations fed back to ILA, exposing it to the closed-loop effects of its own corrections.

\subsection{Learning Reference Adaptation from Observation History}
Similar postures can occur during yielding, holding, or recovery; observation history provides temporal context to distinguish these responses. ILA infers corrections from this history and the nominal reference, without modified references, dedicated interaction sensors, or wrench estimation.

A privileged teacher learns the interaction representation; a student infers it from deployable inputs. A shared decoder maps these representations to corrections:
\begin{equation}
\begin{aligned}
 z_T&=E_T(h_t,r_t,\xi_t^{\mathrm{priv}}),\\
 z_S&=E_S(h_t,r_t),\\
 \Delta r_t&=D(h_t,r_t,z,p,i).
\end{aligned}
\end{equation}
The teacher additionally receives $\xi_t^{\mathrm{priv}}$: nominal and modified arm-reference windows, their residual, the active mask, and its temporal occupancy within the reference window.

The latent $z$ also predicts phase $p$ and per-arm intensity $i$, which condition the decoder. Four phase classes distinguish no interaction, yielding, holding, and recovery; two sigmoid intensity outputs specify correction magnitude for each arm, supervised by normalized per-arm residual amplitudes.

Training couples correction accuracy, motion continuity and recovery, and student inference through
\begin{equation}
\begin{split}
 \mathcal L={}&\lambda_{\mathrm{dec}}\mathcal L_{\mathrm{dec}}
 +\lambda_z\mathcal L_{\mathrm{distill}}+\lambda_p\mathcal L_{\mathrm{phase}}\\
 &+\lambda_i\mathcal L_{\mathrm{intensity}}
 +\lambda_{tr}\mathcal L_{\mathrm{transition}}.
\end{split}
\label{eq:loss}
\end{equation}
$\mathcal L_{\mathrm{dec}}$ combines phase-weighted residual regression and student--teacher trajectory agreement. Its velocity, smoothness, keypoint, and amplitude terms preserve temporal and geometric structure. Zero-correction labels outside interaction and return supervision over recovery windows encourage nominal-reference restoration; the latter penalizes joint, velocity, and keypoint errors over the recovery window and the modified reference's decaying segment. $\mathcal L_{\mathrm{transition}}$ regularizes latent evolution and consecutive residual predictions. $\mathcal L_{\mathrm{distill}}$ aligns teacher and student latents with Huber and cosine terms, while phase and intensity losses supervise the respective auxiliary prediction heads.

The teacher first learns with oracle phase and intensity. Student training aligns its latent with the stop-gradient teacher latent while progressively replacing oracle context with predictions. Deployment retains the student encoder, phase and intensity heads, and decoder to update corrections from the latest observation and action history.
}

\section{Experiments}
\ilarevision{
We examine whether ILA adds compliant guidance while preserving nominal-motion execution, whether the same adapter transfers across trackers without finetuning, and whether reference adaptation enables existing trackers to accommodate human guidance while maintaining whole-body coordination in real-world interactions. Controlled analyses examine how reference supervision and interaction history support these interaction capabilities.
}

\subsection{Experimental Protocol}
\ilarevision{
\textbf{Setting.} An external force guides one hand toward a shoulder-relative target while the robot executes its nominal motion. Two motions, both arms, 81 feasible targets, and three force limits give 243 rollouts per method, with matched guidance conditions across methods.

\textbf{Comparisons.} Standard MotionTracking (MT-Std) is the source tracker; MT-Comp is its native compliance configuration. Baselines include learned compliance from GentleHumanoid~\cite{lu2025gentle} and our CHIP reimplementation (CHIP-R)~\cite{chen2025chip}, and wrench-estimation-based admittance control from MCC~\cite{shi2026mcc}. MimicLite and Sonic test cross-tracker reuse. Full-policy finetuning (FT) adapts each tracker directly; every +ILA configuration keeps the tracker fixed.

\textbf{Evaluation.} Hold-phase hand-to-target distance and positive guidance-force work measure guidance. Joint root-mean-square error (RMSE) measures inactive-arm deviation, post-release whole-body recovery, and no-force tracking against the nominal reference. Maximum horizontal root displacement and fall rate assess whole-body behavior. No-force comparisons use matched settings within each tracker. The appendix provides implementation and measurement details.
}

\begin{table*}[t]
\caption{Matched SR-AOW v1 comparison on G1 over identical Dance1+Walk1 conditions. CHIP-R denotes our reimplementation.}
\label{tab:main}
\WideTableStyle
\setlength{\tabcolsep}{5.5pt}
\begin{tabular*}{0.96\textwidth}{@{\extracolsep{\fill}}lrrrrrr@{}}
\toprule
Method & Target $\downarrow$ & Work $\downarrow$ & Inactive $\downarrow$ & Root $\downarrow$ & Recovery $\downarrow$ & Fall $\downarrow$ \\
 & (cm) & (J) & ($^\circ$) & (m) & ($^\circ$) & (\%) \\
\midrule
GentleHumanoid~\cite{lu2025gentle} & 2.569 & 7.461 & 10.580 & 1.420 & 6.611 & 2.88 \\
MT-Comp~\cite{axellwppr_motiontracking} & 7.079 & 9.677 & 3.206 & 0.993 & 0.641 & 0.00 \\
\textbf{ILA (Ours)} & \textbf{5.684} & 10.859 & 2.815 & 1.246 & 0.619 & 0.00 \\
MCC~\cite{shi2026mcc} & 11.214 & 13.435 & 3.028 & 1.573 & 1.335 & 4.53 \\
CHIP-R~\cite{chen2025chip} & 9.851 & 13.384 & 6.750 & 1.074 & 0.587 & 0.00 \\
MT-Std~\cite{axellwppr_motiontracking} & 17.801 & 13.675 & 2.749 & 1.642 & 1.122 & 2.88 \\
\bottomrule
\end{tabular*}
\end{table*}

\subsection{Policy-Preserving Compliance Retrofit}
\label{sec:retrofit}
\begin{figure*}[t]
\centering
\includegraphics[width=0.96\textwidth]{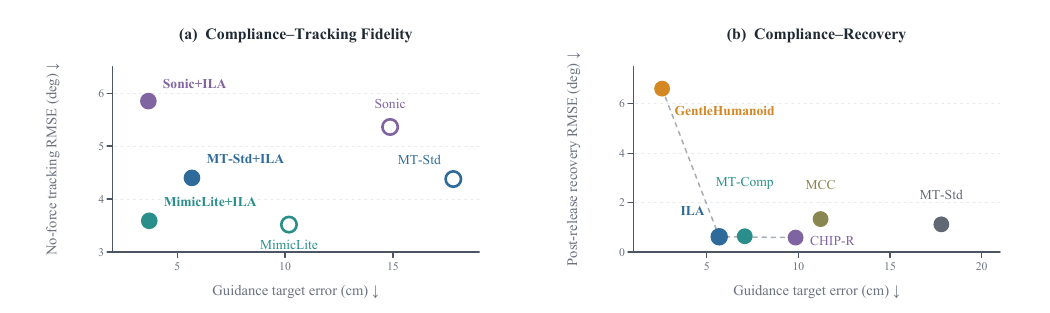}
\caption{Guidance and motion preservation; lower is better on both axes. (a)~Guidance target error versus no-force joint tracking RMSE against each run's original reference. Hollow/filled markers denote each frozen tracker without/with the same ILA; native startup schedules differ across trackers. (b)~Guidance target error versus whole-body post-release recovery under matched SR-AOW v1 conditions on G1; the dashed line shows the non-dominated frontier.}
\label{fig:tradeoff}
\end{figure*}

\ilarevision{
We first test whether reference adaptation adds interaction capability to a fixed tracker. Adding ILA to the source tracker reduces target error by \TargetReductionPct, alongside lower work, root displacement, and fall rate (Table~\ref{tab:main}).

The response also preserves motion selectively: inactive-arm error remains close to its original level, recovery improves after release, and no-force tracking error increases by only 0.5\% (Fig.~\ref{fig:tradeoff}). Paired no-force executions remain close to unadapted trajectories in the five-motion check as well.

Comparisons with existing compliance methods clarify the resulting balance between guidance and motion retention. ILA combines lower target and inactive-arm errors than MCC, CHIP-R, and native MotionTracking compliance, and improves recovery over native compliance. GentleHumanoid reaches closer to the target, while ILA keeps the uninvolved arm closer to its nominal motion and achieves more accurate whole-body recovery after release.
}

\subsection{Cross-Tracker Transfer without Finetuning}
\label{sec:transfer}
\begin{figure}[t]
\centering
\includegraphics[width=\columnwidth]{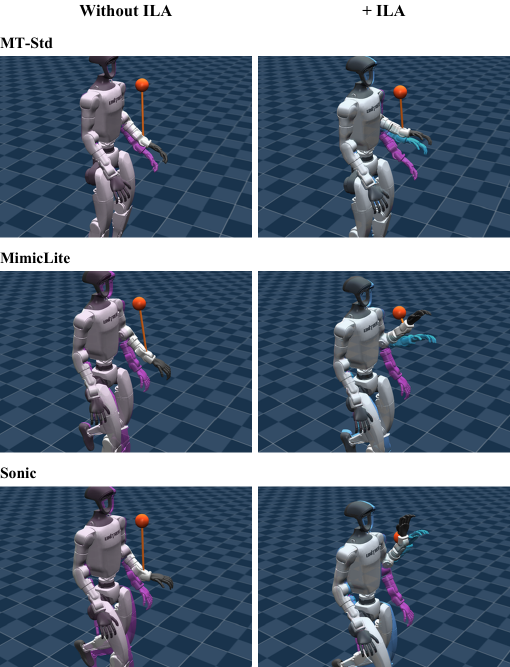}
\caption{Cross-tracker reference adaptation under the same Walk1-left-20~N guidance condition. The thin orange wrist-to-sphere line indicates the applied external guidance force toward the orange target. Magenta denotes the nominal reference. With ILA, interaction-conditioned corrections shift the reference to the cyan pose, allowing the frozen tracker to yield toward the target. Columns compare each tracker without and with the same MT-Std-trained ILA; both adapter and tracker parameters remain fixed.}
\label{fig:transfer}
\end{figure}

\ilarevision{
We test reuse by connecting the same MT-Std-trained adapter to MimicLite and Sonic through deterministic observation and reference mappings, with all adapter and tracker parameters fixed (Table~\ref{tab:transfer}, Fig.~\ref{fig:transfer}). This tests whether one learned interaction module can serve different independently trained humanoid motion controllers.

Target error decreases by \MLReductionPct~on MimicLite and \SonicReductionPct~on Sonic. Both trackers also exhibit lower interaction work, root displacement, and fall rate, with improved whole-body recovery after force release.

Full-policy finetuning provides a second comparison: adapting each target controller directly instead of reusing ILA. Across all three trackers, ILA produces stronger guidance and better post-release recovery than the evaluated finetuning runs. Its no-force tracking-error increase is smaller on MT-Std and MimicLite; Sonic exhibits a larger increase together with substantially stronger guidance (Fig.~\ref{fig:tradeoff}a).
}

\begin{table*}[t]
\caption{Compliance retrofit and cross-tracker reuse. FT finetunes the full target tracking policy. Every +ILA row uses the same frozen MT-Std-trained adapter with the tracker fixed. Guidance uses matched SR-AOW v1 conditions; no-force tracking uses matched settings within each tracker. MT-Std and MT-Std+ILA guidance values follow Table~\ref{tab:main}. Bold denotes the minimum within each tracker group.}
\label{tab:transfer}
\WideTableStyle
\setlength{\tabcolsep}{4pt}
\begin{tabular*}{0.96\textwidth}{@{\extracolsep{\fill}}lrrrrrrr@{}}
\toprule
Method & Target & Work & Inactive & Root XY & Recovery & Fall & No-force tracking \\
 & (cm) & (J) & ($^\circ$) & (m) & ($^\circ$) & (\%) & ($^\circ$) \\
\midrule
MT-Std~\cite{axellwppr_motiontracking} & 17.801 & 13.675 & \textbf{2.749} & 1.642 & 1.122 & 2.88 & \textbf{4.379} \\
MT-FT & 12.780 & 12.004 & 2.836 & 1.358 & 0.808 & 2.47 & 4.624 \\
MT-Std+ILA (Ours) & \textbf{5.684} & \textbf{10.859} & 2.815 & \textbf{1.246} & \textbf{0.619} & \textbf{0.00} & 4.400 \\
\midrule
MimicLite~\cite{mimiclite2026} & 10.182 & 9.495 & 3.276 & 1.440 & 2.409 & 5.76 & \textbf{3.519} \\
MimicLite-FT & 9.856 & 9.251 & 3.349 & 1.493 & 1.931 & 5.35 & 3.724 \\
MimicLite+ILA (Ours) & \textbf{3.705} & \textbf{6.461} & \textbf{2.333} & \textbf{0.787} & \textbf{1.138} & \textbf{0.82} & 3.591 \\
\midrule
Sonic~\cite{luo2026sonic} & 14.871 & 12.274 & \textbf{5.366} & 1.631 & 3.535 & 6.17 & \textbf{5.362} \\
Sonic-FT & 12.793 & 11.950 & 5.648 & 1.721 & 3.026 & 5.89 & 5.663 \\
Sonic+ILA (Ours) & \textbf{3.663} & \textbf{7.195} & 5.460 & \textbf{0.789} & \textbf{1.637} & \textbf{2.06} & 5.849 \\
\bottomrule
\end{tabular*}
\end{table*}

\subsection{Extending Interaction Capability on Real Hardware}
\label{sec:real-world}
\begin{figure}[t]
\centering
\input{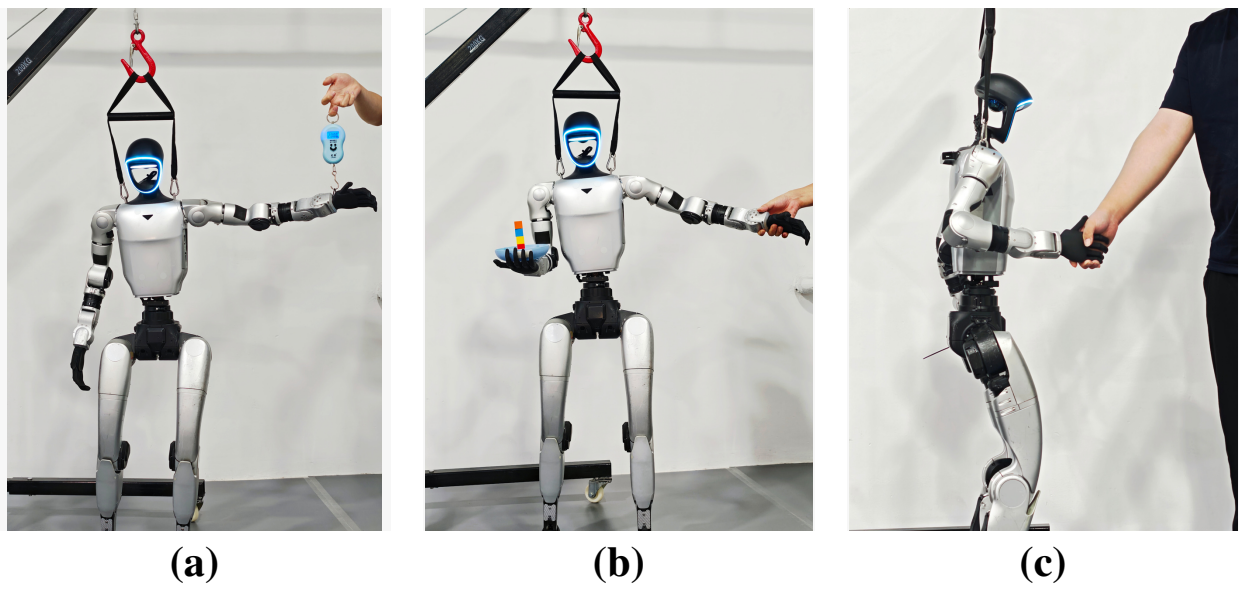}
\caption{Real-world interaction tasks. (a)~Force-gauged wrist pull to a prescribed target. (b)~Unilateral disturbance while the other hand maintains object support. (c)~Human-led handshake. Table~\ref{tab:real-world} reports task metrics.}
\label{fig:real-world}
\end{figure}

\ilarevision{
G1 experiments examine capability extension, concurrent task execution, and whole-body coordination. The same adapter is deployed with all three trackers; comparisons include their original versions, native MotionTracking compliance, and GentleHumanoid. Each method receives ten physical trials per task, counting falls and safety-related terminations as failures (Table~\ref{tab:real-world}, Fig.~\ref{fig:real-world}).

\textbf{Extending interaction capability.} Pulling the wrist toward a lateral-raise target directly tests whether the fixed controller can accommodate an externally guided motion. None of the original trackers reaches the target, whereas every ILA-equipped tracker reaches it in all ten trials.

\textbf{Selective response with concurrent execution.} Unilateral disturbance tests whether the robot can yield with one arm while continuing to support an object with the other. All ILA-equipped trackers maintain support while accommodating the disturbance. This demonstrates selective adaptation: the guided motion changes while the existing controller sustains a separate task during the interaction.

\textbf{Coordination during continuous guidance.} Human-led handshakes require repeated guided motion while maintaining whole-body balance. All ILA-equipped trackers complete the task in every trial, showing that repeated reference adjustments can work together with the tracker's ongoing execution of coordinated whole-body motion.

\textbf{Deployment cost.} ILA takes 0.51~ms for standalone inference on the G1 controller and runs in a separate asynchronous trajectory-generation process, with a small inference cost relative to the 20~ms control cycle.
}

\begin{table}[t]
\caption{Real-world interaction results over ten physical trials per method and task. Methods and sources follow Tables~\ref{tab:main} and~\ref{tab:transfer}.}
\label{tab:real-world}
\CompactTableStyle
\setlength{\tabcolsep}{2pt}
\begin{tabular}{@{}lrrr@{}}
\toprule
Method & \shortstack{Pull force $\downarrow$\\(N)} & \shortstack{Disturbance\\success $\uparrow$} & \shortstack{Handshake\\success $\uparrow$} \\
\midrule
MT-Std & \HWMTPullBase & \HWMTBaseDisturb & \HWMTBaseHandshake \\
MT-Std + ILA (Ours) & \HWMTPullILA & \textbf{\HWMTILADisturb} & \textbf{\HWMTILAHandshake} \\
\addlinespace[2pt]
MimicLite & \HWMLPullBase & \HWMLBaseDisturb & \HWMLBaseHandshake \\
MimicLite + ILA (Ours) & \textbf{\HWMLPullILA} & \textbf{\HWMLILADisturb} & \textbf{\HWMLILAHandshake} \\
\addlinespace[2pt]
Sonic & \HWSonicPullBase & \HWSonicBaseDisturb & \HWSonicBaseHandshake \\
Sonic + ILA (Ours) & \HWSonicPullILA & \textbf{\HWSonicILADisturb} & \textbf{\HWSonicILAHandshake} \\
\midrule
MT-Comp & \HWMTCompPull & \textbf{\HWMTCompDisturb} & \textbf{\HWMTCompHandshake} \\
GentleHumanoid & \HWGentlePull & \textbf{\HWGentleDisturb} & \textbf{\HWGentleHandshake} \\
\bottomrule
\end{tabular}
\par\vspace{1mm}
\begin{minipage}{\columnwidth}\scriptsize
Pull entries give mean force at the lateral-raise target (10/10 reached). NR@$x$ denotes 0/10 reached; $x$ is the mean peak force. Binary entries give successes in ten trials.
\end{minipage}
\end{table}

\subsection{Learning an Interaction-Conditioned Reference Adapter}
\begin{figure}[t]
\centering
\includegraphics[width=\columnwidth]{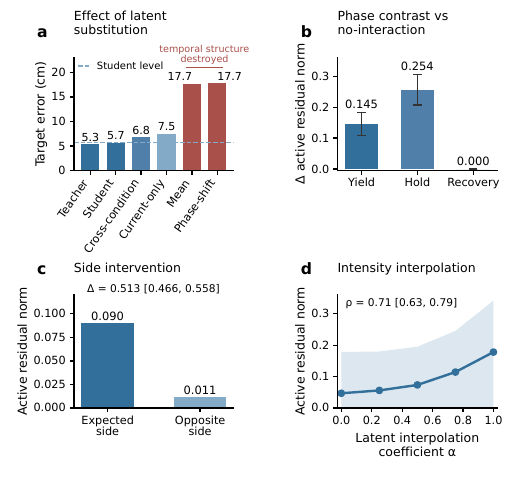}
\caption{Controlled interventions on the latent $z$. (a) Latent substitutions assess the role of interaction context in guidance. (b) Phase-dependent residual activation. (c) Side-dependent adaptation. (d) Intensity-dependent residual magnitude. Reported intervals are 95\% cluster bootstrap CIs (10{,}000 resamples; (a) $n=243$ target clusters across evaluation seeds 101--103, aggregated over three force levels; (b) $n=48$ clusters, (c) $n=117$ mirror pairs, (d) $n=243$ conditions); the shaded band in (d) is the 10--90\% range across all evaluated interaction conditions.}
\label{fig:rai}
\end{figure}

\ilarevision{
We analyze how the adapter learns effective reference changes and uses interaction context to select them.

\subsubsection{Learning from Motion-Reference Targets}
\label{sec:supervision}
We compare paired-reference supervision with two geometric alternatives: a fixed mapping from hand displacement to joint corrections and a local inverse-kinematics solve. The nominal motions, output horizon, active mask, and training procedure are shared, isolating the source of the supervised reference change (Table~\ref{tab:supervision}). Paired-reference supervision produces lower target error, interaction work, and fall rate across the evaluated seeds and motions. These results support learning corrections from temporally structured motion references as an effective way to couple guidance with executable motion.

\subsubsection{Using the Ongoing Interaction}
\label{sec:intent}
To test whether guidance depends on context aligned with the current interaction, we replace the student's latent representation while keeping the decoder, tracker, and external conditions fixed (Fig.~\ref{fig:rai}a). The student approaches the privileged teacher's guidance accuracy. Replacing its representation with the training-set teacher mean or shifting its phase reduces guidance to approximately the unadapted level; using another interaction's representation or only the current observation retains a partial response. These degradations show that effective corrections use the evolving interaction context.

\subsubsection{Selecting When, Where, and How Much to Adapt}
Fixed-state interventions further examine what the learned representation controls (Fig.~\ref{fig:rai}b--d). Holding the observed history, nominal reference, and decoder fixed, we vary only the latent representation. Corrections strengthen during yielding and holding, while the recovery contrast is close to the no-interaction level. Substituting the interaction side redirects the correction, and interpolating latent intensity increases its magnitude. These interventions link the learned context to when, where, and how much the reference changes.
}

\begin{table}[t]
\caption{Residual-supervision sources on MT-Std checkpoint 0509 (separate from the main benchmark). Values are mean $\pm$ standard deviation across three seeds and five motions.}
\label{tab:supervision}
\CompactTableStyle
\setlength{\tabcolsep}{1pt}
\begin{tabular}{lccccc}
\toprule
Source & Target (cm) & Work (J) & Inactive ($^\circ$) & Root (m) & Fall (\%) \\
\midrule
Direct & 14.83$\pm$0.08 & 16.13$\pm$0.18 & 3.03$\pm$0.02 & 2.08$\pm$0.03 & 7.4$\pm$0.1 \\
IK & 11.86$\pm$0.12 & 14.70$\pm$0.14 & 3.18$\pm$0.13 & 1.76$\pm$0.05 & 5.7$\pm$0.2 \\
\shortstack[l]{\textbf{Paired-reference}\\(Ours)} & 7.99$\pm$0.32 & 12.68$\pm$0.62 & 2.68$\pm$0.05 & 1.50$\pm$0.10 & 3.8$\pm$0.4 \\
\bottomrule
\end{tabular}
\end{table}

\subsection{Consistency across Seeds and Motions}
\label{sec:robustness}
\ilarevision{
Across five seeds and five motions on MT-Std, target error varies little with training seed (coefficient of variation \StabCV). Motion-dependent differences are larger; FightAndSports1 has the highest fall rate and recovery error.
}

\ilarevision{
Across simulation and hardware experiments, ILA adds compliant guidance to fixed motion trackers while largely retaining nominal-motion execution. The same adapter transfers across independently trained trackers without finetuning and supports human guidance, concurrent tasks, and whole-body coordination on hardware. Controlled analyses show that paired-reference supervision and evolving interaction history support effective reference corrections, while evaluations across motions and training seeds provide evidence of reproducibility. Together, these findings support compliance as a reusable extension of existing motion tracking.
}

\section{Discussion}
The reference interface provides a useful division of responsibility: ILA learns how interaction should change the motion intent, while the pretrained policy executes the adapted motion. The simulation and hardware evidence connects this architectural separation to selective guidance and retained motion behavior. Learning reference corrections builds on tracker coordination and stabilization skills.

Cross-tracker reuse further shows that the learned interaction response can be separated from the particular controller used during training. Interface mappings preserve the meaning of the adapter's inputs and corrections, while the target tracker determines their realized whole-body response. This explains why deployment can reuse the same parameters yet produce tracker-dependent changes in nominal motion. The resulting capability is reusable across the tested controllers, with guidance and motion retention assessed per tracker.

\section{Conclusion}
We presented ILA, an Interface-Level Adapter that learns interaction-dependent reference corrections while keeping pretrained tracking policies unchanged. By separating reference adaptation from whole-body execution, ILA provides a reusable reference interface for adding compliant interaction to existing whole-body motion controllers.

\appendix[Implementation and Evaluation Details]
\label{app:experiment-details}
\ilarevision{
\textbf{Guidance protocol.} SR-AOW v1 anchors targets to the nominal shoulder at 3~s. A mirrored Fibonacci sphere, inverse-kinematics and collision checks, and three reachable-radius fractions (0.4/0.7/0.9) define the targets. A PD force ($k_p=800$~N/m, $k_d=15$~N$\cdot$s/m) is capped at 20/40/60~N. Each rollout lasts 12~s at 50~Hz: nominal motion, force ramp, hold, and recovery each occupy 3~s. Dance1 and Walk1 were randomly selected before evaluation and fixed across methods. Evaluation seed 101 gives four motion--arm anchors with six to seven directions each, totaling 81 targets across the three reachable-radius fractions.

\textbf{Measurement details.} Distances and positive work use the shoulder anchor frame. No-force RMSE covers all 29 joints over the full rollout, averaged across anchors; paired execution deviation uses the unadapted tracker. We retain the original startup references of each tracker, including small MT-Std interpolation differences.

\textbf{Models and training.} We use the released MimicLite-huge policy and IsaacLab Sonic. GentleHumanoid uses its released checkpoint and official defaults; open-source MCC is adapted to the G1 stack; CHIP-R implements hindsight-perturbation training. Full-policy finetuning uses three seeds per tracker without ILA. The deployed ILA student uses training seed 0, a 64-D latent, and a 32-step reference horizon at 50~Hz. Training uses 1024 environments: 20k teacher steps with oracle context, then 20k and 8k student steps with oracle probability annealed from 1 to 0.5 to 0. The supervision comparison uses three seeds and five motions with MotionTracking checkpoint 0509; the main benchmark uses 0527. Interface mappings align observations and references by joint name, coordinates, and timing.

\textbf{Physical trials.} A force gauge pulls the wrist to a shoulder-height lateral-raise target; Table~\ref{tab:real-world} reports acquisition force for successful reaches and peak force for failures. Handshake success requires three complete cycles between relaxed lower and forearm-horizontal upper boundaries within 3~s, without visible stumbling or a recovery step. ONNX models run on the G1 aarch64 ARM CPU using onnxruntime 1.19.2; standalone batch-1 inference takes 0.51~ms for ILA and 0.75~ms for the frozen motion tracker.
}

\bibliographystyle{IEEEtran}
\bibliography{references}

\begin{thebibliography}{10}
\providecommand{\url}[1]{#1}
\csname url@samestyle\endcsname
\providecommand{\newblock}{\relax}
\providecommand{\bibinfo}[2]{#2}
\providecommand{\BIBentrySTDinterwordspacing}{\spaceskip=0pt\relax}
\providecommand{\BIBentryALTinterwordstretchfactor}{4}
\providecommand{\BIBentryALTinterwordspacing}{\spaceskip=\fontdimen2\font plus
\BIBentryALTinterwordstretchfactor\fontdimen3\font minus \fontdimen4\font\relax}
\providecommand{\BIBforeignlanguage}[2]{{%
\expandafter\ifx\csname l@#1\endcsname\relax
\typeout{** WARNING: IEEEtran.bst: No hyphenation pattern has been}%
\typeout{** loaded for the language `#1'. Using the pattern for}%
\typeout{** the default language instead.}%
\else
\language=\csname l@#1\endcsname
\fi
#2}}
\providecommand{\BIBdecl}{\relax}
\BIBdecl

\bibitem{luo2026sonic}
Z.~Luo, Y.~Yuan, T.~Wang, C.~Li, S.~Chen, F.~Casta\~neda, Z.-A. Cao, J.~Li, D.~Minor, Q.~Ben \emph{et~al.}, ``{SONIC}: Supersizing motion tracking for natural humanoid whole-body control,'' \emph{Science Robotics}, vol.~11, no. 117, p. eaed4592, 2026.

\bibitem{chen2025gmt}
\BIBentryALTinterwordspacing
Z.~Chen, M.~Ji, X.~Cheng, X.~Peng, X.~B. Peng, and X.~Wang, ``Gmt: General motion tracking for humanoid whole-body control,'' \emph{arXiv preprint arXiv:2506.14770}, 2025. [Online]. Available: \url{https://arxiv.org/abs/2506.14770}
\BIBentrySTDinterwordspacing

\bibitem{he2024h2o}
\BIBentryALTinterwordspacing
T.~He, Z.~Luo, W.~Xiao, C.~Zhang, K.~Kitani, C.~Liu, and G.~Shi, ``Learning human-to-humanoid real-time whole-body teleoperation,'' \emph{arXiv preprint arXiv:2403.04436}, 2024. [Online]. Available: \url{https://arxiv.org/abs/2403.04436}
\BIBentrySTDinterwordspacing

\bibitem{he2024omnih2o}
\BIBentryALTinterwordspacing
T.~He, Z.~Luo, X.~He, W.~Xiao, C.~Zhang, W.~Zhang, K.~Kitani, C.~Liu, and G.~Shi, ``Omnih2o: Universal and dexterous human-to-humanoid whole-body teleoperation and learning,'' \emph{arXiv preprint arXiv:2406.08858}, 2024. [Online]. Available: \url{https://arxiv.org/abs/2406.08858}
\BIBentrySTDinterwordspacing

\bibitem{lu2025gentle}
\BIBentryALTinterwordspacing
Q.~Lu, Y.~Feng, B.~Shi, M.~Piseno, Z.~Bao, and C.~K. Liu, ``Gentlehumanoid: Learning upper-body compliance for contact-rich human and object interaction,'' \emph{arXiv preprint arXiv:2511.04679}, 2025. [Online]. Available: \url{https://arxiv.org/abs/2511.04679}
\BIBentrySTDinterwordspacing

\bibitem{margolis2025softmimic}
\BIBentryALTinterwordspacing
G.~B. Margolis, M.~Wang, N.~Fey, and P.~Agrawal, ``Softmimic: Learning compliant whole-body control from examples,'' \emph{arXiv preprint arXiv:2510.17792}, 2025. [Online]. Available: \url{https://arxiv.org/abs/2510.17792}
\BIBentrySTDinterwordspacing

\bibitem{chen2025chip}
\BIBentryALTinterwordspacing
S.~Chen, Z.-A. Cao, Z.~Luo, F.~Casta\~neda, C.~Li, T.~Wang, Y.~Yuan, L.~Fan, C.~K. Liu, and Y.~Zhu, ``Chip: Adaptive compliance for humanoid control through hindsight perturbation,'' \emph{arXiv preprint arXiv:2512.14689}, 2025. [Online]. Available: \url{https://arxiv.org/abs/2512.14689}
\BIBentrySTDinterwordspacing

\bibitem{shi2026mcc}
\BIBentryALTinterwordspacing
H.~Shi, S.~Hu, Y.~Hou, W.~Wang, K.~Liu, and S.~Song, ``Minimalist compliance control,'' \emph{arXiv preprint arXiv:2603.00913}, 2026. [Online]. Available: \url{https://arxiv.org/abs/2603.00913}
\BIBentrySTDinterwordspacing

\bibitem{peng2018deepmimic}
\BIBentryALTinterwordspacing
X.~B. Peng, P.~Abbeel, S.~Levine, and M.~van~de Panne, ``Deepmimic: Example-guided deep reinforcement learning of physics-based character skills,'' in \emph{ACM Transactions on Graphics (SIGGRAPH)}, 2018. [Online]. Available: \url{https://arxiv.org/abs/1804.02717}
\BIBentrySTDinterwordspacing

\bibitem{mahmood2019amass}
\BIBentryALTinterwordspacing
N.~Mahmood, N.~Ghorbani, N.~F. Troje, G.~Pons-Moll, and M.~J. Black, ``Amass: Archive of motion capture as surface shapes,'' in \emph{IEEE International Conference on Computer Vision}, 2019. [Online]. Available: \url{https://arxiv.org/abs/1904.03278}
\BIBentrySTDinterwordspacing

\bibitem{luo2023phc}
\BIBentryALTinterwordspacing
Z.~Luo, J.~Cao, A.~Winkler, K.~Kitani, and W.~Xu, ``Perpetual humanoid control for real-time simulated avatars,'' \emph{arXiv preprint arXiv:2305.06456}, 2023. [Online]. Available: \url{https://arxiv.org/abs/2305.06456}
\BIBentrySTDinterwordspacing

\bibitem{luo2023pacer}
\BIBentryALTinterwordspacing
Z.~Luo, J.~Cao, J.~Merel, A.~Winkler, J.~Huang, K.~Kitani, and W.~Xu, ``Universal humanoid motion representations for physics-based control,'' \emph{arXiv preprint arXiv:2310.04582}, 2023. [Online]. Available: \url{https://arxiv.org/abs/2310.04582}
\BIBentrySTDinterwordspacing

\bibitem{tessler2024maskedmimic}
\BIBentryALTinterwordspacing
C.~Tessler, Y.~Guo, O.~Nabati, G.~Chechik, and X.~B. Peng, ``{MaskedMimic}: Unified physics-based character control through masked motion inpainting,'' \emph{ACM Transactions on Graphics}, 2024. [Online]. Available: \url{https://research.nvidia.com/labs/par/maskedmimic/}
\BIBentrySTDinterwordspacing

\bibitem{he2024hover}
\BIBentryALTinterwordspacing
T.~He, W.~Xiao, T.~Lin, Z.~Luo, Z.~Xu, Z.~Jiang, J.~Kautz, C.~Liu, G.~Shi, X.~Wang, L.~Fan, and Y.~Zhu, ``Hover: Versatile neural whole-body controller for humanoid robots,'' \emph{arXiv preprint arXiv:2410.21229}, 2024. [Online]. Available: \url{https://arxiv.org/abs/2410.21229}
\BIBentrySTDinterwordspacing

\bibitem{he2025asap}
\BIBentryALTinterwordspacing
T.~He, J.~Gao, W.~Xiao, Y.~Zhang, Z.~Wang, J.~Wang, Z.~Luo, G.~He, N.~Sobanbab, C.~Pan \emph{et~al.}, ``Asap: Aligning simulation and real-world physics for learning agile humanoid whole-body skills,'' \emph{arXiv preprint arXiv:2502.01143}, 2025. [Online]. Available: \url{https://arxiv.org/abs/2502.01143}
\BIBentrySTDinterwordspacing

\bibitem{gu2024humanoidgym}
\BIBentryALTinterwordspacing
X.~Gu, Y.-J. Wang, and J.~Chen, ``Humanoid-gym: Reinforcement learning for humanoid robot with zero-shot sim2real transfer,'' \emph{arXiv preprint arXiv:2404.05695}, 2024. [Online]. Available: \url{https://arxiv.org/abs/2404.05695}
\BIBentrySTDinterwordspacing

\bibitem{fu2025humanplus}
\BIBentryALTinterwordspacing
Z.~Fu, Q.~Zhao, Q.~Wu, G.~Wetzstein, and C.~Finn, ``{HumanPlus}: Humanoid shadowing and imitation from humans,'' in \emph{Conference on Robot Learning}, ser. Proceedings of Machine Learning Research, vol. 270, 2025, pp. 2828--2844. [Online]. Available: \url{https://proceedings.mlr.press/v270/fu25a.html}
\BIBentrySTDinterwordspacing

\bibitem{ji2024exbody2}
\BIBentryALTinterwordspacing
M.~Ji, X.~Peng, F.~Liu, J.~Li, G.~Yang, X.~Cheng, and X.~Wang, ``{ExBody2}: Advanced expressive humanoid whole-body control,'' \emph{arXiv preprint arXiv:2412.13196}, 2024. [Online]. Available: \url{https://arxiv.org/abs/2412.13196}
\BIBentrySTDinterwordspacing

\bibitem{liao2025beyondmimic}
\BIBentryALTinterwordspacing
Q.~Liao, T.~E. Truong, X.~Huang, Y.~Gao, G.~Tevet, K.~Sreenath, and C.~K. Liu, ``Beyondmimic: From motion tracking to versatile humanoid control via guided diffusion,'' \emph{arXiv preprint arXiv:2508.08241}, 2025. [Online]. Available: \url{https://arxiv.org/abs/2508.08241}
\BIBentrySTDinterwordspacing

\bibitem{hogan1985impedance}
N.~Hogan, ``Impedance control: An approach to manipulation: Part i---theory,'' \emph{Journal of Dynamic Systems, Measurement, and Control}, vol. 107, no.~1, pp. 1--7, 1985.

\bibitem{ott2015cartesian}
C.~Ott, A.~Albu-Sch\"affer, A.~Kugi, and G.~Hirzinger, ``Cartesian impedance control of redundant and flexible-joint robots,'' \emph{Springer Tracts in Advanced Robotics}, 2008.

\bibitem{ott2010unified}
C.~Ott, R.~Mukherjee, and Y.~Nakamura, ``Unified impedance and admittance control,'' in \emph{IEEE International Conference on Robotics and Automation}, 2010, pp. 554--561.

\bibitem{ott2008passivity}
C.~Ott, A.~Albu-Sch\"affer, A.~Kugi, and G.~Hirzinger, ``On the passivity-based impedance control of flexible joint robots,'' \emph{IEEE Transactions on Robotics}, vol.~24, no.~2, pp. 416--429, 2008.

\bibitem{xu2025facet}
B.~Xu, H.~Weng, Q.~Lu, Y.~Gao, and H.~Xu, ``{FACET}: Force-adaptive control via impedance reference tracking for legged robots,'' in \emph{Conference on Robot Learning}, ser. Proceedings of Machine Learning Research, vol. 305, 2025, pp. 705--720.

\bibitem{zhi2025unifp}
P.~Zhi, P.~Li, J.~Yin, B.~Jia, and S.~Huang, ``Learning a unified policy for position and force control in legged loco-manipulation,'' in \emph{Conference on Robot Learning}, ser. Proceedings of Machine Learning Research, vol. 305, 2025, pp. 652--669.

\bibitem{zhang2026falcon}
Y.~Zhang, Y.~Yuan, P.~Gurunath, I.~Gupta, S.~Omidshafiei, A.-a. Agha-mohammadi, M.~Vazquez-Chanlatte, L.~Pedersen, T.~He, and G.~Shi, ``{FALCON}: Learning force-adaptive humanoid loco-manipulation,'' in \emph{Learning for Dynamics and Control}, ser. Proceedings of Machine Learning Research, vol. 331, 2026, pp. 265--281.

\bibitem{axellwppr_motiontracking}
Axellwppr and contributors, ``{MotionTracking}: Humanoid motion tracking and compliance training,'' \url{https://github.com/Axellwppr/motion_tracking}, 2025, open-source repository.

\bibitem{liu2026lac}
\BIBentryALTinterwordspacing
Y.~Liu, Z.~Gu, W.~Zhu, and M.~Hayashibe, ``Lac: Linear and angular compliance for humanoid whole-body control,'' \emph{arXiv preprint arXiv:2608.25405}, 2026. [Online]. Available: \url{https://arxiv.org/abs/2608.25405}
\BIBentrySTDinterwordspacing

\bibitem{luo2026ceer}
\BIBentryALTinterwordspacing
X.~Luo, X.~Chen, X.~Yin, H.~Wu, B.~Xia, Z.~Chen, J.~Li, B.~Chen, and X.~Cheng, ``Ceer: Compliant end-effector and root control as a unified interface for hierarchical humanoid loco-manipulation,'' \emph{arXiv preprint arXiv:2605.19981}, 2026. [Online]. Available: \url{https://arxiv.org/abs/2605.19981}
\BIBentrySTDinterwordspacing

\bibitem{chen2026dwp}
\BIBentryALTinterwordspacing
Z.~Chen, R.~Li, and Z.~Wang, ``Diversity-aware weight perturbation promotes robust adaptation,'' in \emph{International Conference on Machine Learning}, 2026. [Online]. Available: \url{https://openreview.net/forum?id=5VoK9nUVSD}
\BIBentrySTDinterwordspacing

\bibitem{miki2022perceptive}
T.~Miki, J.~Lee, J.~Hwangbo, L.~Wellhausen, V.~Koltun, and M.~Hutter, ``Learning robust perceptive locomotion for quadrupedal robots in the wild,'' \emph{Science Robotics}, vol.~7, no.~62, p. eabk2822, 2022.

\bibitem{kumar2021rma}
\BIBentryALTinterwordspacing
A.~Kumar, Z.~Fu, D.~Pathak, and J.~Malik, ``Rma: Rapid motor adaptation for legged robots,'' \emph{Robotics: Science and Systems}, 2021. [Online]. Available: \url{https://arxiv.org/abs/2107.04034}
\BIBentrySTDinterwordspacing

\bibitem{kumar2022arma}
\BIBentryALTinterwordspacing
A.~Kumar, Z.~Li, J.~Zeng, D.~Pathak, K.~Sreenath, and J.~Malik, ``Adapting rapid motor adaptation for bipedal robots,'' in \emph{IEEE/RSJ International Conference on Intelligent Robots and Systems}, 2022. [Online]. Available: \url{https://arxiv.org/abs/2205.15299}
\BIBentrySTDinterwordspacing

\bibitem{johannink2019residual}
T.~Johannink, S.~Bahl, A.~Nair, J.~Luo, A.~Kumar, M.~Loskyll, J.~A. Ojea, E.~Solowjow, and S.~Levine, ``Residual reinforcement learning for robot control,'' in \emph{IEEE International Conference on Robotics and Automation}, 2019, pp. 6023--6029.

\bibitem{silver2018residual}
\BIBentryALTinterwordspacing
T.~Silver, K.~Allen, J.~Tenenbaum, and L.~Kaelbling, ``Residual policy learning,'' \emph{arXiv preprint arXiv:1812.06298}, 2018. [Online]. Available: \url{https://arxiv.org/abs/1812.06298}
\BIBentrySTDinterwordspacing

\bibitem{zhao2025resmimic}
\BIBentryALTinterwordspacing
S.~Zhao, Y.~Ze, Y.~Wang, C.~K. Liu, P.~Abbeel, G.~Shi, and R.~Duan, ``{ResMimic}: From general motion tracking to humanoid whole-body loco-manipulation via residual learning,'' \emph{arXiv preprint arXiv:2510.05070}, 2025. [Online]. Available: \url{https://arxiv.org/abs/2510.05070}
\BIBentrySTDinterwordspacing

\bibitem{kang2026splitadapter}
\BIBentryALTinterwordspacing
J.~Kang, H.~Cho, S.~Kang, and D.~Koo, ``Splitadapter: Load-aware humanoid loco-manipulation via factorized adaptation,'' \emph{arXiv preprint arXiv:2606.03297}, 2026. [Online]. Available: \url{https://arxiv.org/abs/2606.03297}
\BIBentrySTDinterwordspacing

\bibitem{zhang2025any2track}
\BIBentryALTinterwordspacing
Z.~Zhang, J.~Guo, C.~Chen, J.~Wang, C.~Lin, Y.~Lian, H.~Xue, Z.~Wang, M.~Liu, H.~Liu, H.~Wang, and L.~Yi, ``Track any motions under any disturbances,'' \emph{arXiv preprint arXiv:2509.13833}, 2025. [Online]. Available: \url{https://arxiv.org/abs/2509.13833}
\BIBentrySTDinterwordspacing

\bibitem{harvey2020robust}
\BIBentryALTinterwordspacing
F.~G. Harvey, M.~Yurick, D.~Nowrouzezahrai, and C.~Pal, ``Robust motion in-betweening,'' \emph{ACM Transactions on Graphics}, vol.~39, no.~4, 2020. [Online]. Available: \url{https://github.com/ubisoft/ubisoft-laforge-animation-dataset}
\BIBentrySTDinterwordspacing

\bibitem{mason2022local}
I.~Mason, S.~Starke, and T.~Komura, ``Real-time style modelling of human locomotion via feature-wise transformations and local motion phases,'' \emph{Proceedings of the ACM on Computer Graphics and Interactive Techniques}, vol.~5, no.~1, 2022.

\bibitem{mimiclite2026}
{RoboParty Lab Team}, ``{MimicLite}: Efficient and effective general humanoid motion tracking,'' \url{https://github.com/EGalahad/mimic-lite}, 2026, technical report: \url{https://github.com/Roboparty/MimicLite/blob/main/mimic-lite.pdf}.

\end{thebibliography}
\end{document}